%% file: main.tex
\documentclass[10pt,twocolumn,letterpaper]{article}

\usepackage{cvpr}              
\input{preamble}
\definecolor{cvprblue}{rgb}{0.21,0.49,0.74}
\usepackage[pagebackref,breaklinks,colorlinks,allcolors=cvprblue]{hyperref}
\usepackage{graphicx}
\usepackage{caption}
\usepackage{lipsum}
\usepackage{amssymb}
\usepackage{tcolorbox}
\usepackage{pifont}
\usepackage[accsupp]{axessibility}

\def\paperID{46745} 
\def\confName{CVPR}
\def\confYear{2026}

\title{SpatialDiff: 3D-Aware Object Movement via Implicit Spatial Modeling}

\author{
    Zheng Liu$^{1*}$ \quad
    Zijian He$^{1}$ \quad
    Huiguo He$^{2}$ \quad
    Weizhi Zhong$^{1}$ \\
    Yejun Tang$^{3}$ \quad
    Huan Yang$^{3}$ \quad
    Kun Gai$^{3}$ \quad
    Guanbin Li$^{1,4,5\dagger}$ \\
    $^{1}$Sun Yat-sen University \qquad
    $^{2}$South China University of Technology \qquad
    $^{3}$Kuaishou Technology \\
    $^{4}$Shenzhen Loop Area Institute \qquad 
    $^{5}$Guangdong Key Laboratory of Big Data Analysis and Processing \\
    {\tt\small liuzh385@mail2.sysu.edu.cn,} {\tt\small liguanbin@mail.sysu.edu.cn}
}

\begin{document}

\twocolumn[{%
\renewcommand\twocolumn[1][]{#1}%
\maketitle
\begin{center}
    \includegraphics[width=1\linewidth]{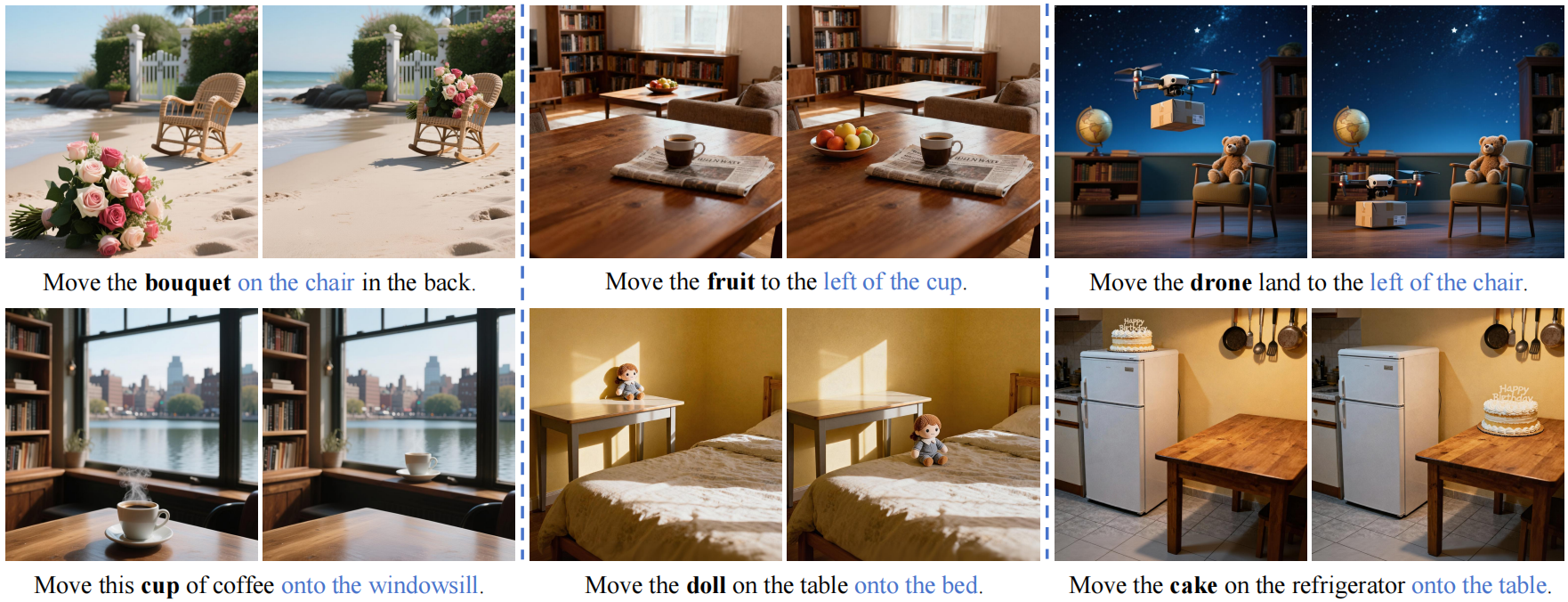}
    \captionof{figure}{Results of our \textbf{Implicit Spatial Modeling} method. Our method captures complex relative positions (e.g., middle, front, top) and performs high-fidelity multi-dimensional transformations while maintaining semantic consistency, making it highly suitable for advanced scene content creation.}
    \label{fig:teaser}
\end{center}
}]

{
    \renewcommand{\thefootnote}{\fnsymbol{footnote}}
    \footnotetext[1]{Work done while interning at Kuaishou Technology.}
    \footnotetext[2]{Corresponding author is Guanbin Li.}
}

\input{sec/0_abstract}    
\input{sec/1_intro}

\input{sec/2_relate}

\input{sec/6_preliminary}

\input{sec/3_method}

\input{sec/4_exp}
\input{sec/5_conclusion}

\clearpage
\section*{Acknowledgments}
This paper is supported by the National Natural Science Foundation of China (No.~62322608), and is also sponsored by CCF-Kuaishou Large Model Explorer Fund (No.~CF-KuaiShou 2024007).

{
    \small
    \bibliographystyle{ieeenat_fullname}
    \bibliography{main}
}

\input{sec/X_suppl}

\end{document}

%% file: sec/0_abstract.tex
\begin{abstract}
Recent advances in image editing allow impressive manipulation of objects, existing methods still struggle to handle spatial movement in complex scenes, such as objects span different depth layers or are partially occluded.
Most image editing methods focus solely on prior information from 2D datasets, emphasizing planar features while lacking support for spatial structures.
Even approaches that incorporate explicit positional information fail to capture true 3D spatial relationships, thus limiting accurate object movement in complex scenes.
In this paper, we present \textbf{SpatialDiff}, a method that effectively captures 3D spatial structures, enabling precise and consistent object movements in complex scenes.
Our core innovations are twofold: (1) \textbf{Implicit 3D Spatial Modeling}, which introduces 3D prior knowledge and enables the model to internally build a comprehensive understanding of the three-dimensional spatial structure; and (2) \textbf{Global Spatial Supervision}, which constrains the latent spatial features to enable the model to perceive changes in object spatial positions caused by editing operations.
Experimental results demonstrate that our method significantly improves the accuracy and fidelity of spatial movement in complex scenes.
\end{abstract}

%% file: sec/1_intro.tex
\section{Introduction}
\label{sec:intro}
\begin{figure} [ht]
    \begin{center}
        \includegraphics[width=\linewidth]{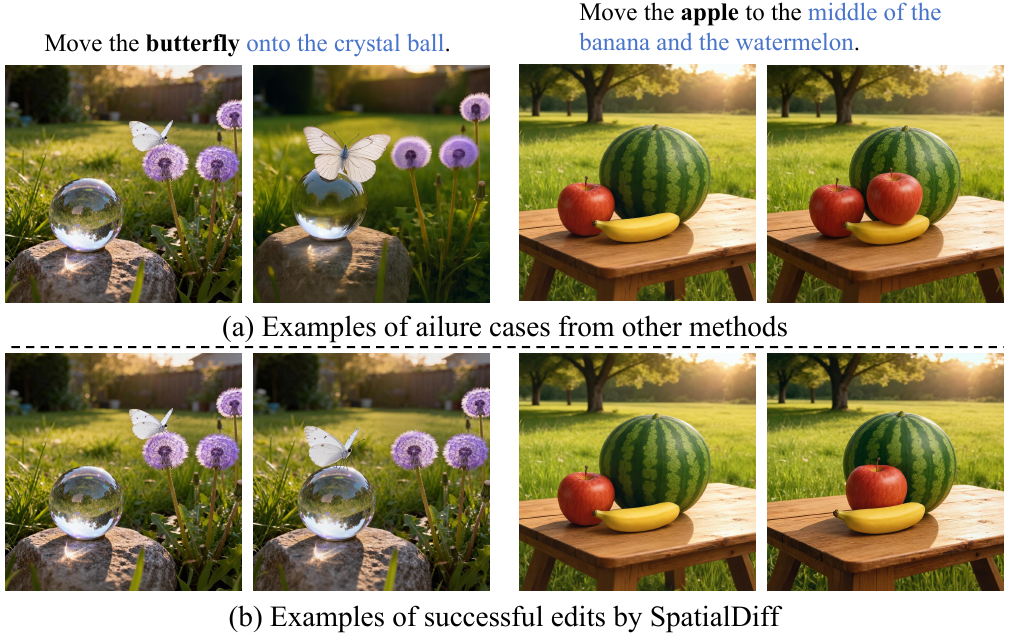} 
    \end{center}
\caption{\textbf{(a)} Examples of failure cases from other methods: (1) fail to maintain natural appearance and 3D consistency; (2) fail to perform object movement correctly in complex scenes. \textbf{(b)} Examples of successful edits by SpatialDiff.}
\label{fig:fail}
\vspace{-10pt}
\end{figure}

Recent years have seen rapid progress in the field of image editing~\cite{zhang2023adding,ci2025describe,wang2024taming,feng2025dit4edit,zhao2024ultraedit,zhou2025fireedit,chen2024anydoor,wang2025omnistyle,zhong2025mod}, especially driven by advances in diffusion-based generative models~\cite{rombach2022high, peebles2023scalable, flux2024, esser2024scaling}. As the quality of base models improves, tasks such as style transfer~\cite{qi2024deadiff,wang2025omnistyle}, object removal and insertion~\cite{chen2024anydoor,alzayer2025magic,li2025blobctrl,huang2025dreamfuse}, and fine-granular content editing~\cite{brooks2023instructpix2pix,cao2023masactrl,huang2024smartedit,zhao2024ultraedit, zhou2025fireedit} have become increasingly viable. 
In many real-world editing scenarios, users may modify an object’s appearance, and relocating it within the scene based on a given instruction is often equally important. However, achieving precise and natural control over object movement within a single image remains challenging, as illustrated in Figure~\ref{fig:fail}.

Methods~\cite{yu2025objectmover, jeon2025crimedit, sun2025smartfreeedit, batifol2025flux, wu2025omnigen2, wu2025qwenimagetechnicalreport, zhang2025unified, liu2025step1x} that operate purely in the 2D domain, particularly those leveraging pretrained diffusion models, provide a more straightforward and widely applicable solution. In these approaches, the editing process remains in image or latent space, the user issues an instruction (e.g., via text), and the model manipulates the source image accordingly. These instruction-driven methods benefit from ease of use and efficiency. Nonetheless, they share a fundamental drawback: in the absence of 3D spatial knowledge (depth, spatial layout), they struggle to guarantee that object motions are consistent in 3D space, especially under user instructions that demand precise spatial relocations. 

A more natural solution to address this is via 3D-aware methods~\cite{bhat2024loosecontrol, yenphraphai2024image, pandey2024diffusion,wang2024diffusion,maillard2025laconic,zhu2025training}. By invoking explicit three-dimensional modelling or reconstruction, such methods aim to reason about object geometry, camera viewpoint, depth, occlusion and spatial relationships. For instance, Diffusion Handles~\cite{pandey2024diffusion} use the estimated depth map to lift diffusion activations for an object to 3D. Diff3DEdit~\cite{wang2024diffusion} predict novel views of the selected object using estimated depth maps, and these novel views act as a geometry critic to correct misalignments in 3D shapes while editing. LACONIC~\cite{maillard2025laconic} leverages the input 3D layout and camera pose to model explicit scene geometry and guide the repositioning of objects.
In principle this grants strong geometric control and spatial consistency. However, when faced with the complex scene editing scenario, these approaches face substantial limitations. 
Explicit 3D reconstruction from a single image is inherently ill-posed: unknown viewpoint, depth ambiguities, occlusion, incomplete geometry and lack of multi-view supervision all degrade the reconstruction quality and thus impair the subsequent editing. 
As a result, though 3D-based pipelines hold promise, their applicability in 2D image editing remains limited.

Summarising the above: on the one hand, 2D-based diffusion editing methods are flexible and effective but lack spatial understanding and thus cannot always execute instruction-level spatial movement accurately; on the other hand, 3D-based editing methods promise spatial reasoning but falter on single-image inputs and complex scene conditions.
This observation raises a key question: 

\textit{(Q) Can we bring the benefits of 3D spatial priors into the 2D image-editing diffusion framework implicitly, rather than through full 3D reconstruction, so that the diffusion transformer (DiT) can understand and control object spatial positioning relations? }

To address this, we propose \textbf{SpatialDiff}, an instruction-driven image editing framework that achieves flexible and precise spatial movement of objects in complex scenes as shown in Figure~\ref{fig:teaser}.
SpatialDiff introduces an \textit{\textbf{Implicit 3D Spatial Modeling}} mechanism by leveraging a 3D visual geometry encoder to extract geometry-aware features from the input image, and a Connector module to align these features with the DiT latent space.
This design enables the model to internalize spatial priors and reason about 3D structure within a purely 2D diffusion process.
Furthermore, to guide the model toward a coherent spatial understanding, we introduce a \textit{\textbf{Global Spatial Supervision}} strategy.
During training, the depth map of the target image is used as an auxiliary supervision signal to regularize the utilization of 3D spatial tokens.
Through this design, our method bridges the gap between 2D and 3D paradigms, mitigating the challenges of single-image 3D reasoning while enriching diffusion-based editing with implicit spatial awareness, enabling precise and spatial consistent object movement. In summary, our key contributions are as follows:

\begin{itemize}
    \item We propose SpatialDiff, the first instruction-driven method that implicitly incorporates 3D prior information to enable precise spatial object movement.
    \item We introduce Global Spatial Supervision, a latent depth supervised training mechanism, which leverages the depth map of the target image during training to smoothly guide the model’s spatial understanding.
    \item Experimental results demonstrate that our model achieves state-of-the-art performance in both instructional movement consistency and image quality preservation.
\end{itemize}

%% file: sec/2_relate.tex
\section{Related Work}
\label{sec:relate}

\subsection{Instruction-based Image Editing}
Instruction-based image editing methods~\cite{kulikov2025flowedit, wu2025chronoedit, huang2024smartedit,fu2023guiding, zhou2025fireedit, labs2025flux1kontextflowmatching, liu2025step1x, zhang2025unified, wu2025qwenimagetechnicalreport} enable flexible image modification by taking natural language commands as input, allowing users to achieve targeted edits on a reference image with improved controllability and a simplified workflow.
FlowEdit~\cite{kulikov2025flowedit} explicitly constructs a transformation path between the reference and target images; ChronoEdit~\cite{wu2025chronoedit} models the image editing process as a continuous video sequence.
Methods such as SmartEdit~\cite{huang2024smartedit}, MGIE~\cite{fu2023guiding}, and FireEdit~\cite{zhou2025fireedit} introduce vision-language models (VLMs) to assist in training diffusion models, thereby enhancing their understanding and reasoning capabilities for complex instructions.
Furthermore, Flux-Kontext~\cite{labs2025flux1kontextflowmatching} significantly improves instruction alignment by scaling up both model parameters and training data.
Step1X-Editing~\cite{liu2025step1x}, BAGEL~\cite{zhang2025unified}, and Qwen-Image-Edit~\cite{wu2025qwenimagetechnicalreport} leverage multimodal large language models (MLLMs) to jointly model input images and user instructions, extracting latent embeddings that are concatenated with Gaussian noise, and then using the model’s in-context learning capability to generate the target image. By incorporating this multimodal understanding, these methods enhance the model’s editing capabilities in complex scenes while maintaining semantic consistency and visual fidelity. However, despite these advances, they remain limited by the lack of injected 3D knowledge, which prevents spatially coherent object motion under fine-grained movement instructions.

\subsection{3D-Aware Image Editing}
Recent works have sought to address spatial inconsistency in image editing by incorporating explicit 3D reasoning into diffusion frameworks~\cite{pandey2024diffusion,wang2024diffusion,maillard2025laconic,zhu2025training,ren2024move,mou2024diffeditor, kumari2024customizing, patashnik2024consolidating}. These approaches introduce geometric awareness through reconstruction or spatial lifting, enabling models to reason about object structure, depth, and viewpoint.
Diffusion Handles~\cite{pandey2024diffusion} lifts diffusion activations into 3D space using estimated depth maps for object manipulation. Diff3DEdit~\cite{wang2024diffusion} synthesizes novel views guided by depth estimation to refine 3D shape alignment during editing. LACONIC~\cite{maillard2025laconic} leverages scene layout and camera pose to model explicit geometry and guide object repositioning. In parallel, single-image 3D reconstruction methods~\cite{liu2023zero,voleti2024sv3d,zhu2025training} attempt to infer object shapes from sparse or single-view inputs, allowing the object to be temporarily “lifted” to 3D, manipulated, and then reprojected back to 2D. 
However, these approaches struggle to handle multiple objects and complex spatial relationships within a single image, where occlusion and layout ambiguity make consistent reconstruction difficult.
Although 3D-aware designs can improve spatial consistency and geometric control, they remain constrained by the inherent challenges of single-image 3D inference. Their reliance on explicit reconstruction, while providing spatial reasoning capabilities, also limits their robustness and scalability in general 2D image editing scenarios.

%% file: sec/6_preliminary.tex
\section{Preliminary}
Diffusion or flow-based models learn data distributions by progressively denoising Gaussian-corrupted samples. Their strong generative ability and controllable sampling make them widely used in instruction-driven image editing.
InstructPix2Pix~\cite{brooks2023instructpix2pix} is a pioneering work in instruction-driven image editing, which fine-tunes a pretrained diffusion model on a large-scale dataset of (input image, instruction, target image) triplets to modify specific objects or regions in an image $x$ according to a user-provided natural language instruction $I$.

In this work, we adopt a flow-based formulation of diffusion, where the forward process linearly interpolates between a clean image and noise to generate intermediate samples:
\begin{equation}
x_t = (1 - t)x + t \epsilon,
\end{equation}
where $\epsilon \sim \mathcal{N}(0, I)$ and $t \in [0, 1]$ denotes the noise level.  
For a flow-based model parameterized by $\theta$, the training objective is defined as:
\begin{equation}
\mathcal{L} = 
\mathbb{E}_{x, t, c, \epsilon \sim \mathcal{N}(0, I)}
\ \| v - v_\theta(x_t, t, c) \|_2^2,
\label{eq:flow_loss}
\end{equation}
where the target velocity is $v = \epsilon - x$. The denoising network $v_\theta$ can be conditioned on auxiliary information $c$, such as a text prompt, a reference image, or other control signals, enabling instruction-guided image editing.

%% file: sec/3_method.tex
\section{Method}
\label{method}

\begin{figure*} [ht]
    \begin{center}
        \includegraphics[width=\linewidth]{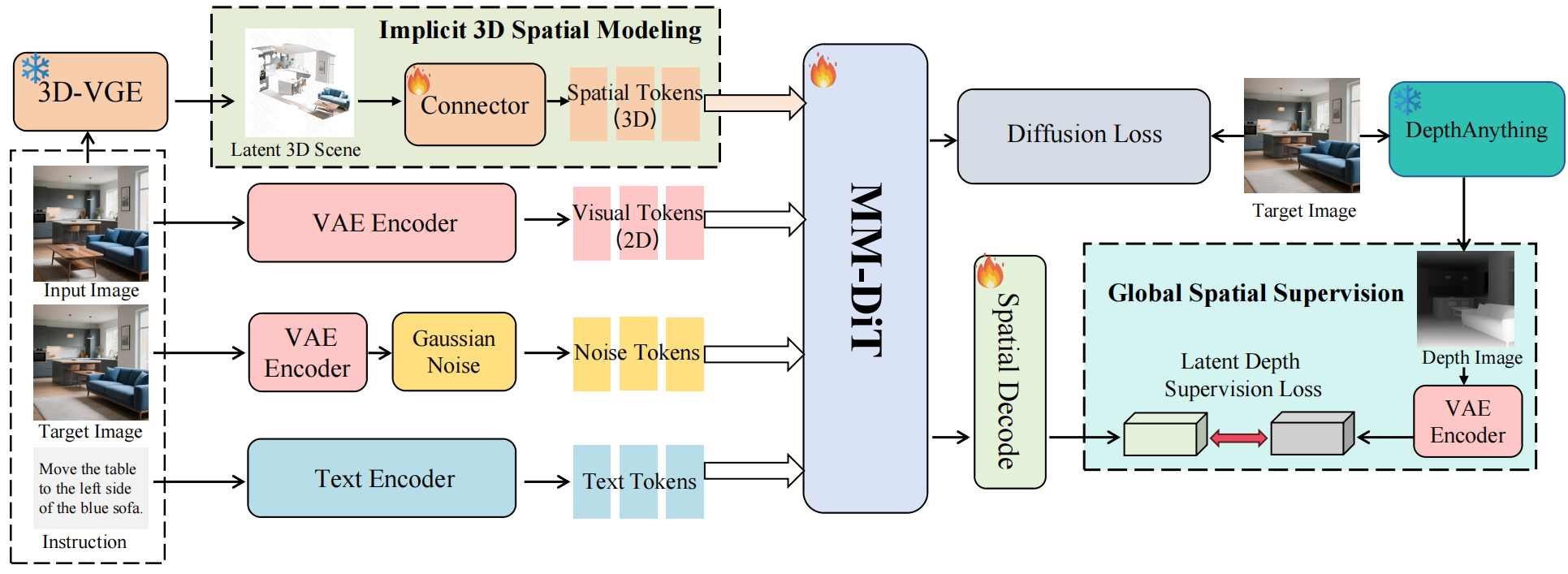} 
    \end{center}
\vspace{-1.5em} 
\caption{The pipeline of SpatialDiff. SpatialDiff leverages a single image along with its latent 3D scene representation to achieve precise control over object positions in 2D image scenes, while preserving the global spatial structure and semantic consistency.}
\label{fig:overview}
\vspace{-1em}
\end{figure*}

We introduce SpatialDiff, an implicit modeling approach that incorporates 3D spatial information from the input image into the image editing process. 
We detail Implicit 3D Spatial Modeling in Section~\ref{method:3dmodeling}, Global Spatial Supervision in Section~\ref{method:depthalignment}, and the two-stage training strategy in Section~\ref{method:strategy}.

\subsection{Implicit 3D Spatial Modeling}
\label{method:3dmodeling}

Existing image editing methods typically handle object movement based on 2D prior knowledge, making them inadequate for complex scenes involve occlusion or depth variation.
We argue that this limitation primarily stems from the model's lack of spatial understanding of the 3D world, and thus propose to incorporate Implicit 3D Spatial Modeling (ISM) into its internal representations.

\paragraph{3D Visual Geometry Encoder.}
To endow the diffusion backbone with a spatially grounded understanding, we extract spatial-aware features from the input image using a 3D Visual Geometry Encoder (3D-VGE). This encoder is based on the Transformer backbone of a recent 3D foundational model, VGGT~\cite{wang2025vggt}, which directly infers all key 3D attributes of a scene, including camera parameters, point maps, depth maps, and 3D point tracks, from one image. 

Given an input image $x_{0}$, we obtain its spatial-aware representation as:
\begin{equation}
S = \mathrm{3D\text{-}VGE}(x_{0}), \quad S \in \mathbb{R}^{l' \times d'},
\label{eq:3dvge}
\end{equation}
where $l'$ denotes the token sequence length and $d'$ the feature dimension of the spatial tokens.
By leveraging the backbone features without using the task-specific prediction heads, 3D-VGE provides implicit 3D geometric priors to guide the image editing process, enabling the model to benefit from 3D spatial knowledge without explicitly reconstructing depth, surface normals, or other 3D attributes. However, since these extracted features differ from the DiT latent representation, an alignment module is required before they can be effectively integrated.

\paragraph{Connector for Spatial Alignment.}
We introduce a Connector module that aligns the 3D Visual Geometry Encoder features with the latent feature space of the diffusion transformer.
The connector employs a set of learnable query tokens $Q_l \in \mathbb{R}^{l \times d}$, where $l$ corresponds to the token sequence length in the DiT space and $d$ denotes the token dimension.
These query tokens selectively extract and align spatial information from the spatial features $S$ through a cross-attention mechanism to match the latent representation of the DiT:
\begin{equation}
\hat{Q}_l = \mathrm{Attn}(Q_l, S) = \mathrm{Softmax}\left(\frac{QK^T}{\sqrt{d}}\right)V,
\label{eq:crossattn}
\end{equation}
where
\begin{equation}
Q = Q_l W_Q^T, \quad K = S W_K^T, \quad V = S W_V^T.
\end{equation}
Here, $W_Q, W_K, W_V$ are learnable projection matrices of dimensions $\mathbb{R}^{d \times d}$, $\mathbb{R}^{d \times d'}$, and $\mathbb{R}^{d \times d'}$, respectively.  
Through this cross-attention alignment, the Connector module distills spatial-aware information into a representation that aligns well with the DiT latent space.

\paragraph{Token Integration.}
The aligned 3D spatial tokens are then concatenated with other conditional tokens to form the complete latent sequence:
\begin{equation}
C = [x, \hat{Q}_l, I],
\label{eq:concat}
\end{equation}
where $x$ denotes the latent tokens of the input image, $Q_l$ denotes the aligned spatial tokens, and $I$ the instruction tokens.  
This unified token sequence is then processed by the multimodal layers of the DiT backbone, where context learning integrates local appearance features, spatial cues, and instructions across the entire sequence. Through this multimodal interaction, SpatialDiff implicitly captures 3D spatial structures, including object depth, relative positioning, and inter-object spatial relationships, enabling precise and geometrically consistent manipulations even in complex scenes with occlusion or varying depth layers.

\subsection{Global Spatial Supervision}
\label{method:depthalignment}
 
Although implicit 3D spatial modeling introduces valuable 3D knowledge, we observe that even when an object is successfully relocated, traces may remain at its original position, or the overall semantic consistency of the scene may be affected (see Figure~\ref{fig:ablation_study}, Model~B).
This suggests that while the aligned 3D spatial features help the model comprehend object geometry and its initial position, they primarily reinforce static spatial memory rather than driving dynamic spatial updates. As a result, the model lacks a mechanism to dynamically guide the scene to remain consistent with the target image during object movement.

To address this issue, we introduce Global Spatial Supervision (GSS), which applies depth-guided supervision \textbf{during training} to align the original 3D spatial tokens with the spatial positions of the target image.
We explored two strategies for depth supervision:  (1) Explicit Depth Supervision, which enforces pixel level consistency between decoded 3D spatial features and target depth maps in image space, and  (2) Latent Depth Supervision, which constrains their correspondence in the VAE latent space. 
While the first strategy, an explicit approach, provides hard alignment, it often leads to unnatural editing results and causes the loss of content in non-edited regions (see Figure~\ref{fig:ablation_study}, Model~C).
Supervision in latent space emphasizes high level spatial and semantic relationships over low level pixel details, delivers a smooth and robust training signal less sensitive to subtle variations in detail, and enables effective integration of 3D spatial priors.
Therefore, we adopt Latent Depth Supervision in this work, which promotes consistency in the latent geometry and better preserves relative spatial relationships among scene elements.

Formally, we obtain the target image’s depth map using a pretrained monocular depth estimator~\cite{yang2024depth}:
\begin{equation}
d_{\text{tgt}} = \mathrm{DepthAnything}(x_{\text{tgt}}),
\label{eq:depthmap}
\end{equation}
where $x_{\text{tgt}}$ denotes the ground-truth edited image in the training pair.

Specifically, during training, we obtain $\hat{s}$, the DiT-processed spatial tokens that encode spatial aware information and map them into the VAE latent space via a learnable spatial decode head $\bar{\mathcal{D}}$.  
We then enforce consistency with the encoded target depth representation using a mean squared error (MSE) loss:
\begin{equation}
\mathcal{L}_{\text{GSS}} =
\|\, \bar{\mathcal{D}}(\hat{s}) -
\mathcal{E}(d_{\text{tgt}}) \,\|_2^2,
\label{eq:gss}
\end{equation}
where $\mathcal{E}(\cdot)$ denotes the encoder of the VAE.  

Unlike prior 3D-aware approaches that rely on explicit reconstruction or depth estimation during inference, our method leverages explicit depth information solely as a smooth auxiliary supervision signal in training.
This design preserves the implicit nature of our spatial modeling while enabling the initial 3D information to be dynamically updated during training, encoding geometry features in a globally consistent manner.
As a result, SpatialDiff achieves a more robust and coherent understanding of spatial structure, effectively suppressing residual artifacts while preserving implicit 3D reasoning without requiring any explicit 3D reconstruction at inference.

\subsection{Training Strategy}
\label{method:strategy}
To ensure stable convergence and fully exploit the capability of the Connector, we employ a two-stage training strategy. In the first stage, only the Connector module is optimized to align 3D spatial features with the DiT latent space, guided by the standard diffusion loss:
\begin{equation}
    \mathcal{L}_{Align} = \mathbb{E}_{x_t, t, C, \epsilon \sim \mathcal{N}(0, I)} \| v - v_\theta(x_t, t, C)\|_2^{2},
\label{eq:stage1_loss}
\end{equation}

In the second stage, we jointly optimize the DiT backbone, the Connector module, and the spatial decoder head that projects the 3D spatial features into the VAE latent space. During this stage, the training objective incorporates Global Spatial Supervision to ensure that the model maintains consistency throughout the generation process:
\begin{equation}
\mathcal{L} = \mathbb{E}_{x_t, t, C, \epsilon \sim \mathcal{N}(0, I)} \| v - v_\theta(x_t, t, C) \|_{2}^{2} + \lambda \cdot \mathcal{L}_{\text{GSS}},
\label{eq:all_loss}
\end{equation}
where $\lambda$ = 0.01 is a weighting factor balancing the flow matching loss and the GSS loss.

%% file: sec/4_exp.tex
\section{Experiments}

\begin{figure*} [ht]
    \begin{center}
        \includegraphics[width=\linewidth]{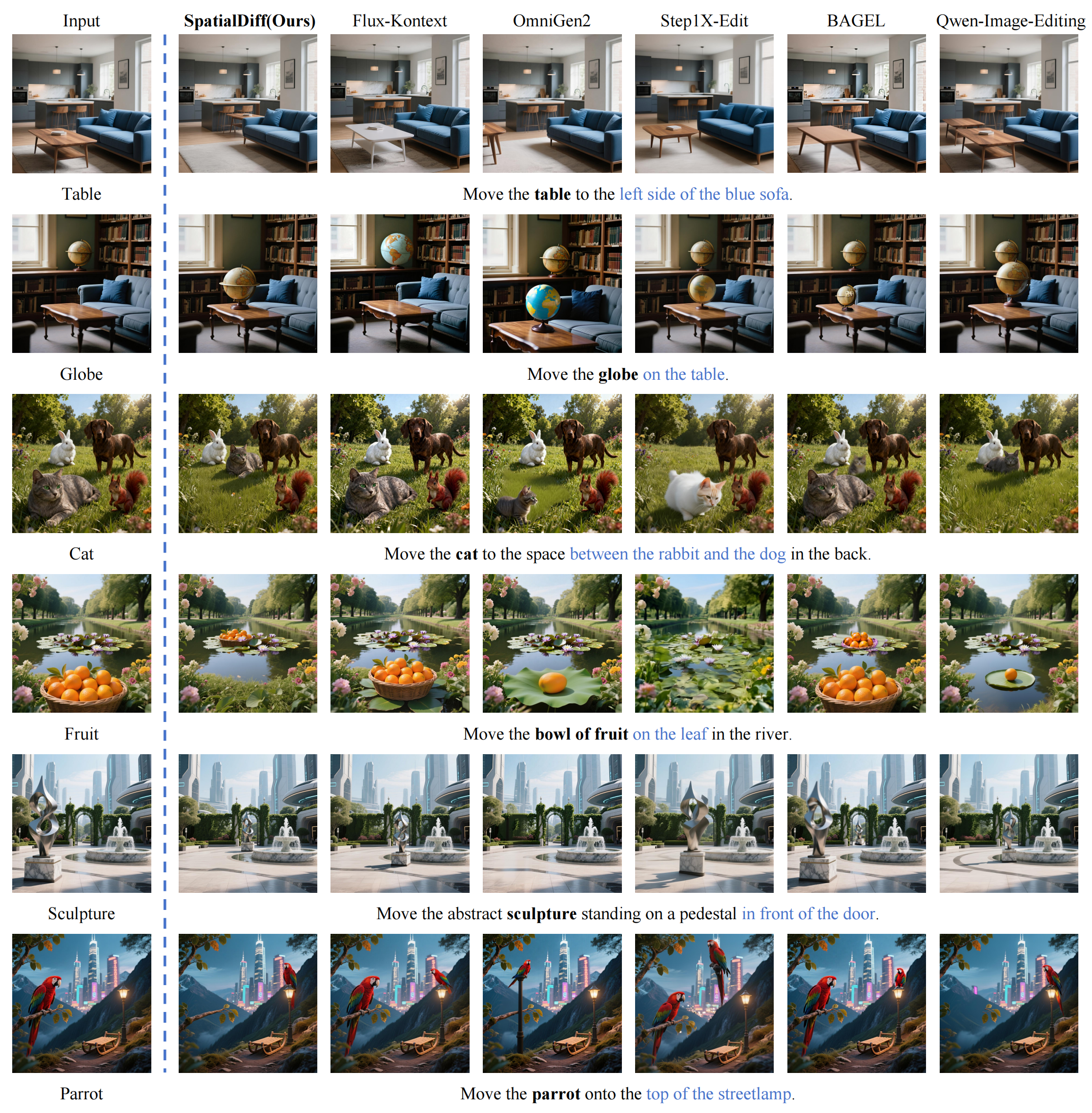} 
    \end{center}
\vspace{-1em} 
\caption{\textbf{Qualitative comparison}. The leftmost image represents the input image, with the target object to be edited shown below it. Each example is accompanied by the corresponding editing instruction, where the blue text indicates the spatial position involved in the editing operation.}
\label{fig:compare}
\end{figure*}

\setlength{\tabcolsep}{3pt}
\begin{table*}[ht]
\centering
\caption{\textbf{Quantitative comparison} on SpatialBench. GPT-SC, GPT-PQ, and GPT-O refer to the metrics evaluated by GPT-5, while Qwen-SC, Qwen-PQ, and Qwen-O refer to the metrics evaluated by Qwen3-VL-32B-Instruct. Scores range from 0 to 1. $\uparrow$: higher is better.}
\vspace{-8pt}
\small
\renewcommand{\arraystretch}{1}%
\begin{tabular*}{1.0\textwidth}{@{\extracolsep{\fill}}lcccccc@{}}
\toprule
Method & GPT-SC$\uparrow$ & GPT-PQ$\uparrow$ & GPT-O$\uparrow$  & Qwen-SC$\uparrow$ & Qwen-PQ$\uparrow$ & Qwen-O$\uparrow$\\
\midrule
Flux-Kontext~\cite{batifol2025flux} & 0.292 & 0.848 & 0.498 & 0.261 & 0.767 & 0.447 \\
OmniGen2~\cite{wu2025omnigen2}  & 0.301 & 0.661 & 0.446 & 0.305 & 0.687 & 0.458\\
Step1X-Edit~\cite{liu2025step1x}   & 0.484 & 0.785 & 0.616 & 0.492 & 0.691 & 0.583\\
BAGEL~\cite{zhang2025unified} & 0.368 & 0.709 & 0.511 & 0.390 & 0.642 & 0.500 \\
Qwen-Image-Editing~\cite{wu2025qwenimagetechnicalreport} & 0.666 & 0.882 & 0.766 & 0.632 & 0.813 & 0.717 \\
\midrule
\textbf{SpatialDiff (Ours)} & \textbf{0.803} & \textbf{0.886} & \textbf{0.843} & \textbf{0.778} & \textbf{0.838} & \textbf{0.807}\\
\bottomrule
\end{tabular*}
\label{tab:quantative_compare}
\end{table*}
\setlength{\tabcolsep}{6pt}

\subsection{Experimental Setups}
\noindent\textbf{Implementation Details.}
We use Flux-Kontext ~\cite{batifol2025flux} as the base model for instruction-driven image editing, and employ VGGT ~\cite{wang2025vggt} as the 3D foundation model to extract 3D features from the reference image. To align the 3D spatial features with the DiT space, we construct the Connector module using 8 cross-attention layers, with the learnable query tokens sequence length set to 1024. The model training is based on the code provided by ~\cite{wu2025less}, with the LoRA rank set to 64. The training images have a resolution of 512 $\times$ 512. In all experiments, we set the learning rate to 1e-4 on 8 GPUs, employed AdamW with a batch size of 4 per GPU, and performed 3k steps of training.

\noindent\textbf{Datasets and Evaluation Benchmarks.}
Due to the lack of datasets for instruction-driven spatial object movement in 2D image editing tasks, we adapted the OBJect-3DIT~\cite{michel2023object} movement dataset. Specifically, we collected the source and target images from OBJect-3DIT and used Qwen3-VL-32B-Instruct~\cite{qwen3technicalreport} to generate editing instructions describing the transformation from source to target images, thereby constructing a movement dataset containing 20k (reference image, instruction, target image) triplets. This dataset was used for training, without relying on any 3D asset information(e.g., coordinates).  To more comprehensively evaluate the model’s performance across different scenarios, we construct an additional complex scene benchmark \textbf{SpatialBench} containing 100 images with foreground, midground, and background objects. The comprehensive evaluation benchmark consists of 50 original OBJect-3DIT test images and 100 additionally constructed images depicting complex real-world scenes. We evaluate the two benchmarks separately, results on SpatialBench are presented in Sections ~\ref{exp:compare_quan} and ~\ref{exp:compare_qual}, while results on the OBJect-3DIT test images are provided in the Supplementary Material.

\noindent\textbf{Comparison Methods.}
We compared SpatialDiff with state-of-the-art instruction-driven image editing models, including Flux-Kontext~\cite{labs2025flux1kontextflowmatching}, Step1X-Edit~\cite{liu2025step1x}, OmniGen2~\cite{wu2025omnigen2}, BAGEL~\cite{zhang2025unified}, and Qwen-Image-Editing~\cite{wu2025qwenimagetechnicalreport}. Most of these methods employ a multimodal large language model (MLLM) to jointly encode the reference image and the editing instruction, enabling them to capture richer spatial position information.

\noindent\textbf{Metrics.}
Following VIEScore~\cite{ku2024viescore}, we adopt three evaluation metrics: SC (Semantic Consistency), PQ (Perceptual Quality), and O (Overall Score). SC measures the consistency between the edited result and the given editing instruction, without considering image quality; PQ evaluates the edited image, including the consistency of the edited object, the stability of unedited regions, and whether the overall image appears realistic and free of artifacts. The overall score O is computed based on SC and PQ, defined as O = (SC $\times$ PQ)$^{\frac{1}{2}}$. In our experiments, both GPT-5~\cite{openai2025} and Qwen3-VL-32B-Instruct~\cite{qwen3technicalreport} were used to score the results according to these criteria.

\subsection{Quantitative Comparison}
\label{exp:compare_quan}
As shown in Table~\ref{tab:quantative_compare}, SpatialDiff achieves the best overall performance across all quantitative metrics for spatial movement task.  
In terms of Semantic Consistency (SC), it attains a GPT-SC score of 0.803 and a Qwen-SC score of 0.778, substantially outperforming existing methods.  This result indicates that SpatialDiff precisely interprets and executes user instructions for spatial movement.  
For Perceptual Quality (PQ), both GPT-PQ (0.886) and Qwen-PQ (0.838) rank highest, reflecting strong preservation of global object consistency and natural visual appearance.  
Consequently, the overall score (O) also reaches the top (GPT-O: 0.843, Qwen-O: 0.807), confirming that SpatialDiff achieves a balanced improvement in both semantic accuracy and visual fidelity. 
In contrast, Flux-Kontext achieved artificially high PQ scores due to minimal image modifications, but their poor SC scores reveal the underlying failure in editing. 
While Qwen-Image-Editing achieves high image quality thanks to its powerful pretraining, its ability to follow instructions remains relatively limited (GPT-SC: 0.666, Qwen-SC: 0.632). 
Other methods (e.g., OmniGen2, Step1X-Edit, and BAGEL) struggle to balance editing fidelity and object consistency, resulting in suboptimal performance in both aspects. Although GPT and Qwen differ in scoring scales, their relative rankings of the methods are highly consistent, further validating the reliability of the results. 
In summary, SpatialDiff achieves state-of-the-art performance in semantic accuracy and visual fidelity.

\subsection{Qualitative Comparison}
\label{exp:compare_qual}
Figure~\ref{fig:compare} presents qualitative results that highlight the precision and stability of our method in performing spatial movement.  
In line 1, our approach successfully relocates the table to the left side of the blue sofa while maintaining its appearance and consistency with the original scene. Our method reasoned in 3D, interpreting “left” relative to the sofa rather than the camera view. Other methods fail to achieve the correct spatial relocation, often introducing structural distortion or inconsistent object shapes. 

Regarding scene coherence and background integrity, our method achieves superior global coordination.  
For line 3, when the cat is moved between the rabbit and the dog, background textures such as grass and shadows remain intact, and spatial relationships among objects are preserved naturally. Unedited regions also remain visually consistent, without redundant texture regeneration or boundary blurring.  
In comparison, OmniGen2 produces noticeable object deformation, Step1X-Edit merges foreground and background content incorrectly, BAGEL leaves ghost artifacts of the original cat, and Qwen-Image-Editing, although placing the cat approximately correctly, removes other unedited objects from the scene.

Our method also demonstrates strong 3D spatial understanding and depth consistency. In line 5, it correctly captures front–back relationships and depth ordering, naturally positioning the sculpture in front of the door.
Other approaches such as Flux-Kontext and Qwen-Image-Editing appear reasonable in 2D projection but fail to maintain depth coherence, leading to weak spatial integration between objects and the door. Overall, qualitative comparisons show that our method excels in superior instruction following, global consistency, and spatial reasoning.

\subsection{Ablations}

\setlength{\tabcolsep}{2pt}
\begin{table*}[h]
\centering
\caption{\textbf{Ablation study} on SpatialBench. \textbf{FT} denotes fine-tuning the attention modules of DiT using LoRA. \textbf{ISM} (Implicit 3D Spatial Modeling) represents introducing 3D tokens from the input image through a 3D foundation model, and aligning them to the DiT feature space. \textbf{EDS} (Explicit Depth Supervision) indicates applying supervision on 3D tokens in the pixel space, while \textbf{LDS} (Latent Depth Supervision) applies supervision in the VAE latent space. GPT-SC, GPT-PQ, and GPT-O refer to the metrics evaluated by GPT-5, while Qwen-SC, Qwen-PQ, and Qwen-O refer to the metrics evaluated by Qwen3-VL-32B-Instruct. Scores range from 0 to 1. $\uparrow$: higher is better.}
\small
\renewcommand{\arraystretch}{1}%
\begin{tabular*}{1.0\textwidth}{@{\extracolsep{\fill}}lcccccccccc@{}}
\toprule
Method & FT & ISM & EDS & LDS & GPT-SC$\uparrow$ & GPT-PQ$\uparrow$ & GPT-O$\uparrow$  & Qwen-SC$\uparrow$ & Qwen-PQ$\uparrow$ & Qwen-O$\uparrow$\\
\midrule
Baseline &  &  &  &  & 0.236 & 0.831 & 0.443 & 0.310 & 0.723 & 0.473\\
Model A  & \checkmark & &  &  & 0.398 & 0.795 & 0.563 & 0.427 & 0.739 & 0.562\\
Model B  & \checkmark & \checkmark &  &  & 0.518 & 0.743 & 0.620 & 0.513 & 0.684 & 0.592\\
Model C  & \checkmark & \checkmark & \checkmark &  & 0.566 & 0.801 & 0.673 & 0.527 & 0.745 & 0.627\\
SpatialDiff & \checkmark & \checkmark &  & \checkmark & \textbf{0.804} & \textbf{0.871} & \textbf{0.837} & \textbf{0.796} & \textbf{0.835} & \textbf{0.815}\\
\bottomrule
\end{tabular*}
\label{tab:ablation_study}
\end{table*}

Table~\ref{tab:ablation_study} presents the ablation quantitative results on SpatialBench, systematically analyzing the contribution of each component in SpatialDiff.

\noindent \textbf{Implicit 3D Spatial Modeling (ISM).}
For a fair comparison, we fine-tune the baseline model via LoRA on the movement dataset (Model A). This results in a noticeable improvement in SC (0.236 → 0.398 on GPT-SC) but a slight decrease in PQ (0.831 → 0.795). This indicates that while fine-tuning brings a limited improvement in the model’s responsiveness to spatial instructions, it may also introduce unnatural distortions, highlighting the trade-off between instruction adherence and image quality. The visual observation (“parrot floating in mid-air”) corroborates this quantitative evidence.
Building upon this, introducing implicit 3D knowledge via a 3D foundation model (Model B) further boosts SC (0.398 → 0.518) and O (0.563 → 0.620), demonstrating that 3D-aware features substantially strengthen spatial reasoning and object localization.

\begin{figure} [h]
    \begin{center}
        \includegraphics[width=\linewidth]{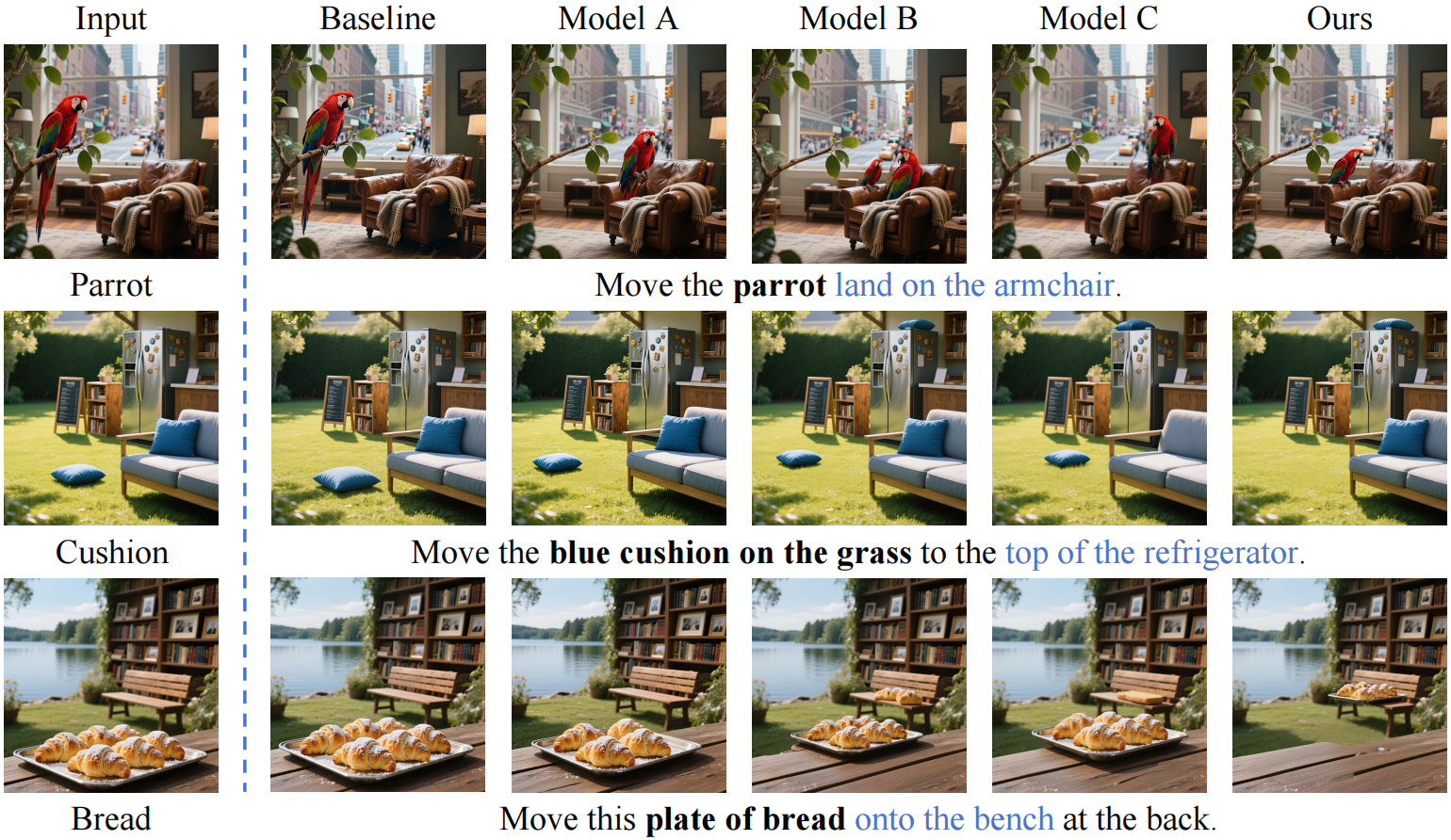} 
    \end{center}
\vspace{-1.5em}
\caption{\textbf{Ablation study}. Model A denotes LoRA fine-tuning applied on the Baseline using the movement dataset; Model B denotes the introduction of Implicit 3D Spatial Modeling only; Model C denotes the incorporation of both Implicit 3D Spatial Modeling and Explicit Depth Supervision. 
}
\label{fig:ablation_study}
\end{figure}

\noindent \textbf{Explicit Depth Supervision (EDS).}
Applying pixel-space depth supervision (Model C) enforces stronger spatial constraints, slightly improving O (0.620 → 0.673 on GPT-O; 0.592 → 0.627 on Qwen-O).
Although this helps align spatial understanding, it causes over-constrained learning that affects non-edited regions (e.g., line 2 in Figure~\ref{fig:ablation_study}, the pillow on sofa should not be removed). Hence, EDS improves structural precision but at the expense of visual stability.

\noindent \textbf{Latent Depth Supervision (LDS).}
When supervision is applied in the latent space (SpatialDiff), the model achieves the best overall balance, with large gains in all metrics (GPT-SC: 0.804, GPT-PQ: 0.871, GPT-O: 0.837). This indicates that latent-level guidance provides smooth, semantically coherent constraints that preserve global scene consistency and high perceptual quality simultaneously.

\setlength{\tabcolsep}{6pt}

\subsection{User Study}
We conducted a user study with 35 participants, asking them to rate each generated image in terms of SC and PQ. Each participant evaluated ten results per method, resulting in 4200 votes. 
The results are reported in Table~\ref{tab:user_study}. Our method consistently achieves higher human-evaluated scores across all three metrics, H-SC, H-PQ, and H-O (overall performance). These results are generally consistent with the quantitative comparisons among the methods. The user study details are provided in supplementary material.

\setlength{\tabcolsep}{3pt}
\begin{table}[ht]
\centering
\caption{\textbf{User Study}. H-SC refers to the assessment of instruction-following ability by humans, H-PQ refers to the evaluation of image quality by humans, and H-O represents the combined metric of both. Scores range from 1 to 5. $\uparrow$: higher is better.}
\vspace{-0.5em} 
\small
\renewcommand{\arraystretch}{1}%
\begin{tabular*}{0.48\textwidth}{@{\extracolsep{\fill}}lccc@{}}
\toprule
Method & H-SC$\uparrow$ & H-PQ$\uparrow$ & H-O$\uparrow$ \\
\midrule
Flux-Kontext~\cite{batifol2025flux} &1.874  &3.433  &2.536  \\
OmniGen2~\cite{wu2025omnigen2}  &2.258  &2.692  &2.465  \\
Step1X-Edit~\cite{liu2025step1x}  &2.271  &2.616  &2.437  \\
BAGEL~\cite{zhang2025unified} &2.759  &3.351  &3.041  \\
Qwen-Image-Editing~\cite{wu2025qwenimagetechnicalreport} &3.513  &3.611  &3.562  \\
\midrule
\textbf{SpatialDiff (Ours)} & \textbf{4.711} & \textbf{4.720} & \textbf{4.715} \\
\bottomrule
\end{tabular*}
\label{tab:user_study}
\end{table}
\setlength{\tabcolsep}{6pt}

%% file: sec/5_conclusion.tex
\vspace{-10pt}
\section{Conclusion}
In this work, we introduced SpatialDiff, a novel instruction-based method designed to enable precise and consistent object movement in complex scenes. Unlike previous 2D editing methods that rely solely on planar priors, SpatialDiff incorporates 3D spatial awareness into the editing process through two key designs: Implicit 3D Spatial Modeling and Global Spatial Supervision.
Implicit 3D Spatial Modeling integrates 3D knowledge into the latent space to enhance the model’s understanding of spatial information, while Global Spatial Supervision provides latent-level constraints on the modeled features, smoothly guiding consistent object movement during training.
Experiments demonstrate that SpatialDiff achieves superior spatial reasoning and perceptual fidelity, producing geometrically coherent results even with occlusions and multi-depth structures.
We believe this study opens new avenues for integrating 3D understanding into diffusion models for spatially-perceptive image editing.

%% file: sec/X_suppl.tex
\clearpage
\setcounter{page}{1}
\setcounter{section}{0}
\maketitlesupplementary

In this supplementary material, we first present quantitative and qualitative results on OBJect-3DIT, comparing SpatialDiff against various baseline methods. This is followed by additional visual results generated by our approach. Furthermore, we provide comparisons with a commercial baseline and present additional ablation studies. Finally, we detail the setup of our user study and conclude with a discussion on the limitations of SpatialDiff alongside potential directions for future work. 

\section{Results on OBJect-3DIT}
\begin{figure*} [tbh]
    \begin{center}
        \includegraphics[width=0.95\linewidth]{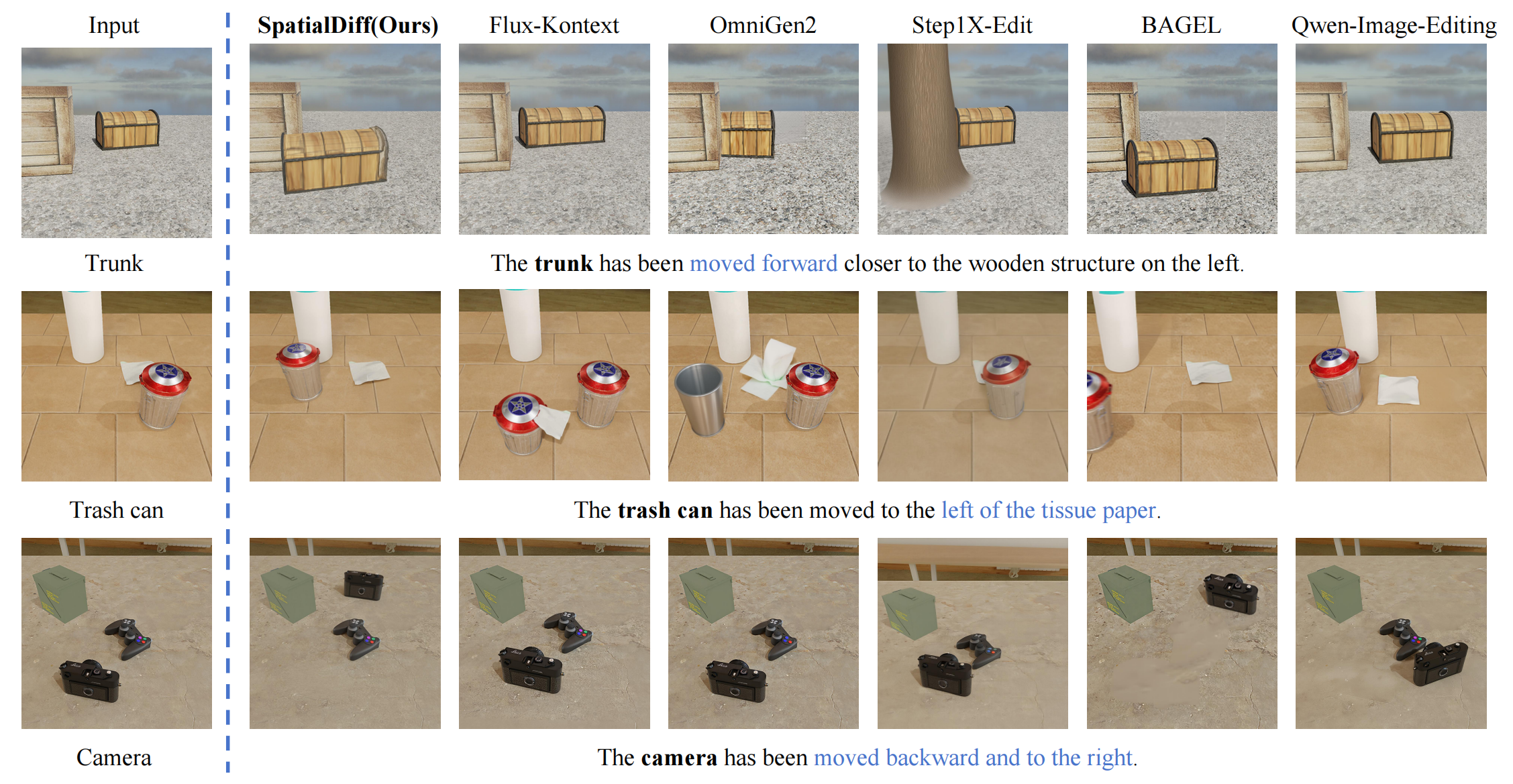} 
    \end{center}
\vspace{-10pt}
\caption{\textbf{Qualitative comparison} on OBJect-3DIT. The leftmost image represents the input image, with the target object to be edited shown below it. Each example is accompanied by the corresponding editing instruction, where the blue text indicates the spatial position involved in the editing operation.}
\label{fig:3DBench}
\end{figure*}

We obtained the quantitative results on the OBJect-3DIT test dataset, as shown in Table~\ref{tab:quantative_compare2}. Our method consistently achieves the best overall performance on this dataset. Notably, OBJect-3DIT features relatively simpler scenes. Compared with the results on SpatialBench (Table~\ref{tab:quantative_compare}), most existing methods perform better in instruction-following under both GPT-SC and Qwen-SC evaluations. The visualization results on this dataset are shown in Figure~\ref{fig:3DBench}.

\section{Additional Qualitative Results}
\begin{figure*}[t]
    \centering 
    \includegraphics[width=0.95\linewidth]{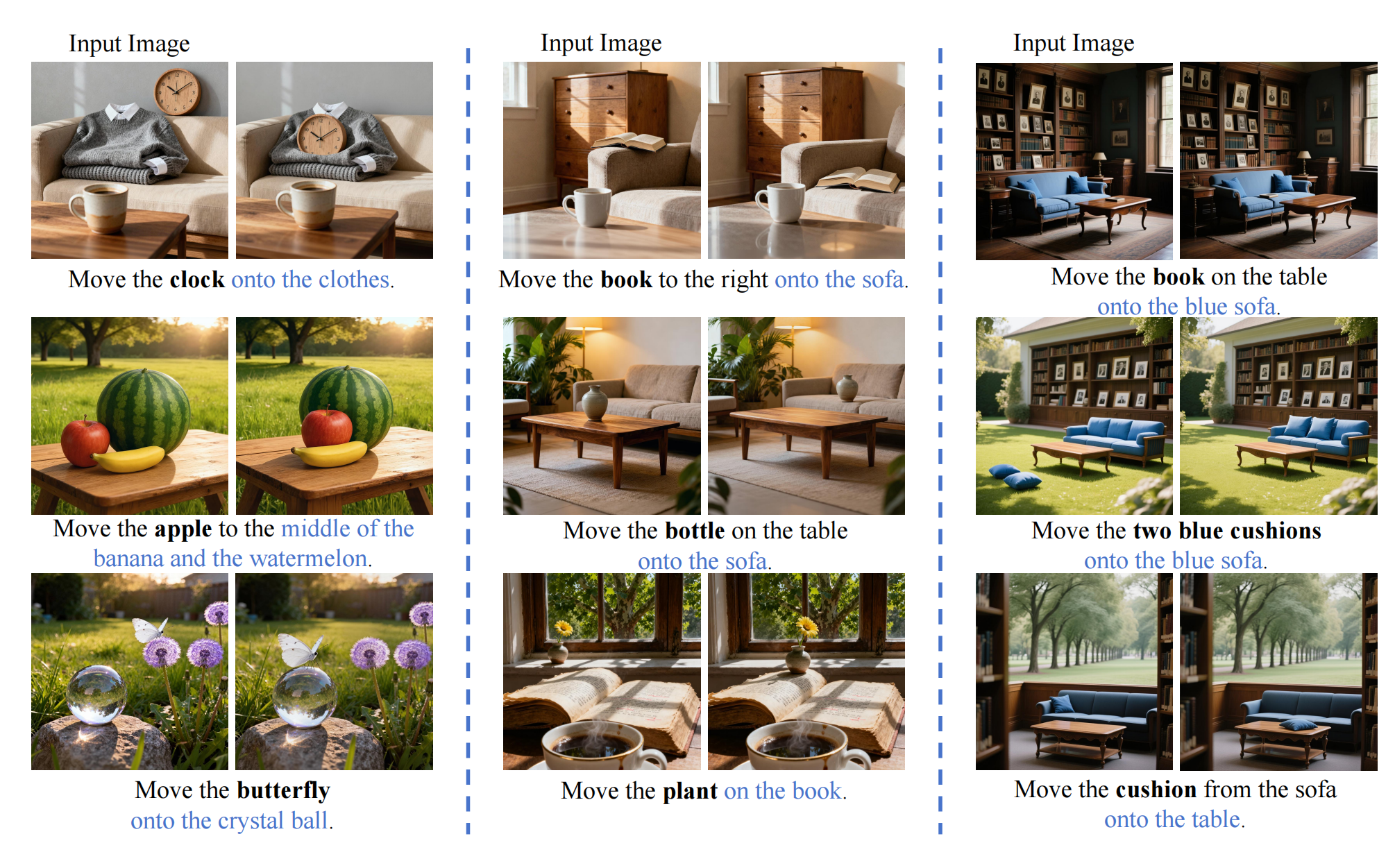}
    \vspace{-10pt}
    \caption{\textbf{More results of SpatialDiff}. Each example is accompanied by the corresponding editing instruction, where the blue text indicates the spatial position involved in the editing operation.}
    \label{fig:more_vis}
\end{figure*}

Figure~\ref{fig:more_vis} shows more results of complex scene object movement. We observe that in the clock example (Row 1, Column 1), SpatialDiff not only moves the clock onto the clothes but also generates a natural indentation, indicating that the movement operation aligns with realistic 3D physical behavior. In the apple example (Row 2, Column 1), the apple is accurately placed between the watermelon and the banana, an instance involving significant front–back occlusion, which demonstrates that SpatialDiff has a strong understanding of 3D spatial relationships. Moreover, in other examples involving depth inconsistencies or occlusions, SpatialDiff is still able to achieve superior performance in both instruction following and object consistency.

\section{Comparison with Commercial Baseline}
Our method continues to outperform the commercial baseline Nano Banana 2 in Table~\ref{tab:commercial}. Specifically, while Nano Banana 2 maintains competitive performance in certain generation quality metrics such as GPT-PQ, SpatialDiff achieves significant margins in spatial instruction adherence (Qwen-SC) and overall accuracy (Qwen-O).

\section{Additional Ablation}
We evaluated the variant, \textit{Full model w/o ISM}, by replacing VGGT features with learnable queries while retaining LDS. Tab.~\ref{tab:ablation_ISM} shows a large drop in instruction following, validating the critical role of ISM. Compared to “w/o LDS”, LDS acts as \emph{linker} that couples depth supervision with VGGT priors, enforcing spatial consistency.

\section{User Study Details}
\begin{figure} [tbh]
    \begin{center}
        \includegraphics[width=\linewidth]{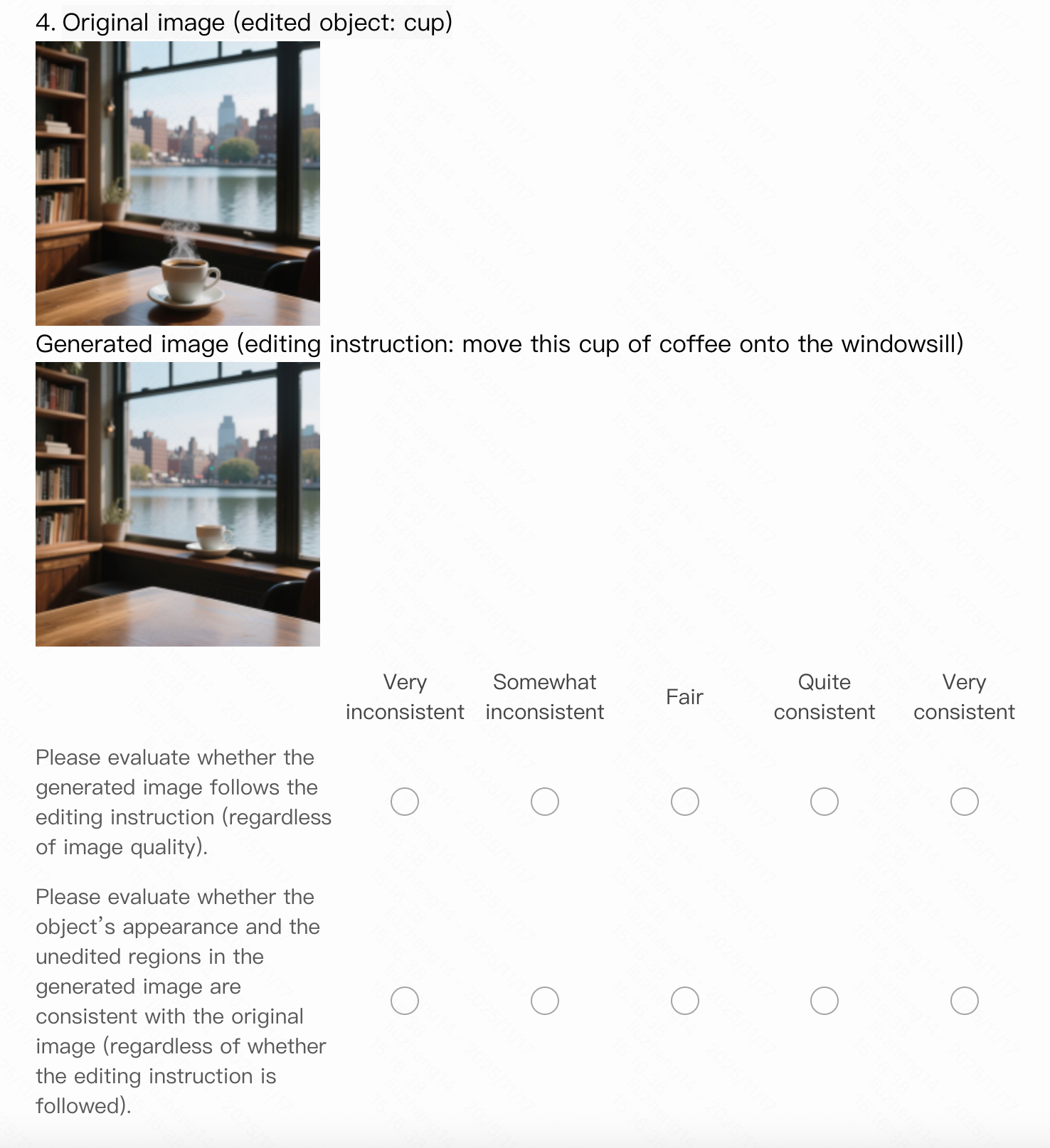} 
    \end{center}
\caption{\textbf{Screenshot of our user study rating interface.}}
\label{fig:user_detail}
\end{figure}

Figure~\ref{fig:user_detail} shows the user study interface for evaluating Semantic Consistency (H-SC) and Perceptual Quality (H-PQ). Each case includes two questions, one on Semantic Consistency and the other on Perceptual Qualit. H-SC primarily evaluates whether the generated image follows the editing instruction, regardless of image quality; H-PQ primarily evaluates whether the attributes and shape of the edited object have changed and whether the unedited regions remain consistent, without considering whether the instruction was completed. Participants rate their responses on a 5-point scale: (1) ``Very inconsistent'', (2) ``Somewhat inconsistent'', (3) ``Fair'', (4) ``Quite consistent'', and (5) ``Very consistent''.

\section{Limitations and Discussion}
\begin{figure}[h]
    \centering
    \includegraphics[width=\linewidth]{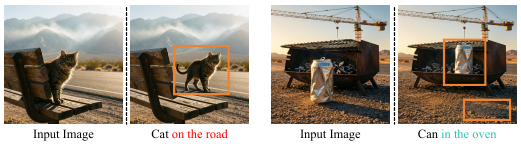}
    \caption{Limitations of the approach.}
    \label{fig:limitation} 
\end{figure}

In Fig.~\ref{fig:limitation}, incorrect perspective scaling and incomplete shadow removal are still observed in SpatialDiff. This reflects a limitation in physical reasoning required to fully handle complex object interactions. Precise distance control also remains challenging due to natural language ambiguity. 

In SpatialDiff, the tokens extracted from the input image by the VGGT backbone are first fused into a learnable query via cross-attention, and then concatenated with other information along the token sequence dimension within MMDiT. While this design introduces useful 3D priors and implicitly models 3D spatial positional information within the network, the additional conditioning tokens increase the computational burden during both training and inference, thereby reducing efficiency. With 1024 spatial tokens, the inference time increased from 29s to 59s. A potential direction for future work is to integrate these 3D priors directly into the model module parameters, so that the extra conditioning computation can be removed during inference—that is, concatenating the 3D prior tokens only during training, while relying solely on the 2D image tokens as conditioning information at inference time.

\setlength{\tabcolsep}{3pt}
\begin{table*}[t]
\centering
\caption{\textbf{Quantitative comparison} on OBJect-3DIT. GPT-SC, GPT-PQ, and GPT-O refer to the metrics evaluated by GPT-5, while Qwen-SC, Qwen-PQ, and Qwen-O refer to the metrics evaluated by Qwen3-VL-32B-Instruct. Scores range from 0 to 1. $\uparrow$: higher is better.}
\small
\renewcommand{\arraystretch}{1}%
\begin{tabular*}{1.0\textwidth}{@{\extracolsep{\fill}}lcccccc@{}}
\toprule
Method & GPT-SC$\uparrow$ & GPT-PQ$\uparrow$ & GPT-O$\uparrow$  & Qwen-SC$\uparrow$ & Qwen-PQ$\uparrow$ & Qwen-O$\uparrow$\\
\midrule
Flux-Kontext~\cite{batifol2025flux} & 0.424 & 0.831 & 0.594 & 0.253 & \textbf{0.858} & 0.466 \\
OmniGen2~\cite{wu2025omnigen2} & 0.416 & 0.587 & 0.494 & 0.407 & 0.691 & 0.530\\
Step1X-Edit~\cite{liu2025step1x} & 0.528 & 0.637 & 0.580 & 0.552 & 0.765 & 0.650\\
BAGEL~\cite{zhang2025unified} & 0.697 & 0.518 & 0.601 & 0.643 & 0.700 & 0.671 \\
Qwen-Image-Editing~\cite{wu2025qwenimagetechnicalreport} & 0.783 & 0.807 & 0.795 & 0.746 & 0.839 & 0.791 \\
\midrule
\textbf{SpatialDiff (Ours)} & \textbf{0.891} & \textbf{0.833} & \textbf{0.862} & \textbf{0.776} & 0.847 & \textbf{0.811}\\
\bottomrule
\end{tabular*}
\label{tab:quantative_compare2}
\end{table*}
\setlength{\tabcolsep}{6pt}

\setlength{\tabcolsep}{3pt}
\begin{table*}[h]
\centering
\caption{Quantitative results compared with a commercial baseline.}
\label{tab:commercial}
\small
\renewcommand{\arraystretch}{1}%
\begin{tabular*}{1.0\textwidth}{@{\extracolsep{\fill}}lcccccc@{}}
\toprule
Method & GPT-SC$\uparrow$ & GPT-PQ$\uparrow$ & GPT-O$\uparrow$ & Qwen-SC$\uparrow$ & Qwen-PQ$\uparrow$ & Qwen-O$\uparrow$ \\
\midrule
Nano Banana 2 & 0.711 & \textbf{0.892} & 0.796 & 0.594 & \textbf{0.840} & 0.706 \\
\textbf{SpatialDiff (Ours)} & \textbf{0.803} & 0.886 & \textbf{0.843} & \textbf{0.778} & 0.838 & \textbf{0.807}\\
\bottomrule
\end{tabular*}
\end{table*}
\setlength{\tabcolsep}{6pt}

\setlength{\tabcolsep}{3pt}
\begin{table*}[h]
\centering
\caption{Additional Ablation Studies on the ISM Module.}
\label{tab:ablation_ISM}
\small
\renewcommand{\arraystretch}{1}%
\begin{tabular*}{1.0\textwidth}{@{\extracolsep{\fill}}lcccccc@{}}
\toprule
Method & GPT-SC$\uparrow$ & GPT-PQ$\uparrow$ & GPT-O$\uparrow$ & Qwen-SC$\uparrow$ & Qwen-PQ$\uparrow$ & Qwen-O$\uparrow$ \\
\midrule
Full model w/o LDS & 0.518 & 0.743 & 0.620 & 0.513 & 0.684 & 0.592\\
Full model w/o ISM & 0.656 & \textbf{0.882} & 0.761 & 0.635 & 0.775 & 0.702 \\
Full model (SpatialDiff) & \textbf{0.804} & 0.871 & \textbf{0.837} & \textbf{0.796} & \textbf{0.835} & \textbf{0.815} \\
\bottomrule
\end{tabular*}
\end{table*}
\setlength{\tabcolsep}{6pt}

\begin{figure*}[ht]
\vspace{-2mm} 
\begin{tcolorbox}[
    colback=gray!5!white,          
    colframe=gray!80!black,        
    boxrule=0.5pt,                 
    arc=3pt,                       
    left=6pt, right=6pt, top=4pt, bottom=4pt, 
    title=\textbf{Prompt for PQ Score}, 
    fonttitle=\small\bfseries,     
    fontupper=\small               
]

\textbf{Task Definition:}\\
You will be provided with two images: one is the reference image before editing, and the other is the result image after editing. The description \texttt{\{prompt\}} specifies a spatial adjustment operation applied to the target object \texttt{\{edit\_object\}} in the reference image. Although the description indicates the intended operation, your evaluation should \textbf{not} focus on whether the editing instruction is successfully completed. Instead, focus on the overall visual quality, naturalness, and spatial consistency of the edited image.
\vspace{1.5mm}

\textbf{Scoring Criteria:}
\begin{itemize}
    \setlength{\itemsep}{0pt}
    \setlength{\parsep}{0pt}
    \item \textbf{Stability:} Whether the non-edited regions remain stable, with continuous background structure, texture, and lighting, free from noticeable damage or blurring.
    \item \textbf{Artifacts:} Whether there are artifacts, ghosting, edge misalignment, or residual traces around the edited object.
    \item \textbf{Consistency:} Whether lighting, shadow, perspective, and depth relationships are consistent and reasonable after editing, ensuring an overall natural and realistic appearance.
\end{itemize}
\vspace{1.5mm}

\textbf{Scoring Range:}
\begin{itemize}
    \setlength{\itemsep}{0pt}
    \setlength{\parsep}{0pt}
    \item \textbf{0.0 - 0.2 (Severe distortion):} The edited area is highly unnatural; non-edited areas are damaged or contain significant artifacts; the overall image looks inconsistent.
    \item \textbf{0.2 - 0.4 (Poor):} Noticeable flaws or artifacts in the edited area; object edges or background appear incoherent; low overall naturalness.
    \item \textbf{0.4 - 0.6 (Fair):} The editing operation is roughly achieved but with some visual inconsistencies such as local blurriness, incorrect shadows, or slight artifacts.
    \item \textbf{0.6 - 0.8 (Good):} The editing appears natural; the object blends well with the background; only minor inconsistencies are present.
    \item \textbf{0.8 - 1.0 (Excellent):} The editing is smooth and seamless; non-edited regions remain completely consistent; the image looks natural, harmonious, and free of visible artifacts or discontinuities.
\end{itemize}
\vspace{1.5mm}

\textbf{Input Format:} You will receive two images: the first is the pre-edit image, and the second is the post-edit image.\\
\textbf{Output Format:} Provide \textbf{only a single numerical score} between 0.0 and 1.0 to assess the overall visual naturalness and consistency.
\end{tcolorbox}
\vspace{-2mm}
\caption{Prompt for PQ Score.}
\label{fig:prompt_pq}
\vspace{-5pt}
\end{figure*}

\begin{figure*}[ht]
\vspace{-2mm}
\begin{tcolorbox}[
    colback=gray!5!white,          
    colframe=gray!80!black,        
    boxrule=0.5pt,                 
    arc=3pt,                       
    left=6pt, right=6pt, top=4pt, bottom=4pt, 
    title=\textbf{Prompt for SC Score}, 
    fonttitle=\small\bfseries,     
    fontupper=\small               
]

\textbf{Task Definition:}\\
You will be provided with two images: one is the reference image before editing, and the other is the result image after editing. The description \texttt{\{prompt\}} specifies a spatial adjustment operation applied to the target object \texttt{\{edit\_object\}} in the reference image. The editing operation may involve movement of the object in two-dimensional directions (up, down, left, right) or in three-dimensional space (near, far, depth). Your task is to evaluate, based on the before-and-after images, whether the editing result accurately follows the operation described in the instruction. \textbf{Do not} evaluate the visual quality of the image; only assess whether the editing operation has been completed as instructed.
\vspace{1.5mm}

\textbf{Scoring Range:}
\begin{itemize}
    \setlength{\itemsep}{0pt}
    \setlength{\parsep}{0pt}
    \item \textbf{0.0 - 0.2 (Not completed at all):} The edited image is almost unrelated to the instruction; the target object did not move as expected, or the direction/position is clearly wrong.
    \item \textbf{0.2 - 0.4 (Partially incorrect):} Some changes are visible, but the movement direction or magnitude does not match the instruction, or the editing effect is only partially reflected.
    \item \textbf{0.4 - 0.6 (Partially correct):} The object roughly moves according to the instruction, but noticeable deviations exist, such as inaccurate position, wrong depth direction, or incorrectly updated background.
    \item \textbf{0.6 - 0.8 (Mostly correct):} The editing result largely matches the instruction; the object’s movement direction and position are generally correct, with only minor errors or unnatural aspects.
    \item \textbf{0.8 - 1.0 (Fully correct):} The edited image accurately executes the spatial movement described in the instruction; direction, distance, and depth relationships are natural and reasonable.
\end{itemize}
\vspace{1.5mm}

\textbf{Input Format:} You will receive two images: the first is the pre-edit image, and the second is the post-edit image.\\
\textbf{Output Format:} Provide \textbf{only a single numerical score} between 0.0 and 1.0 to assess the accuracy of the spatial editing operation.
\end{tcolorbox}
\vspace{-2mm}
\caption{Prompt for SC Score.}
\label{fig:prompt_sc}
\vspace{-5pt}
\end{figure*}

\begin{table*}[b]
    \centering
    \caption{Licenses for released assets}
    \label{tab:licenses}
    \begin{tabular}{ll}
        \toprule
        \textbf{Asset} & \textbf{License} \\
        \midrule
        \midrule
        UNO~\citep{wu2025less}  & Apache license 2.0 \\
        Flxu-Kontext~\citep{batifol2025flux} & Apache license 2.0 \\
        OBJect-3DIT~\cite{michel2023object} & CreativeML Open RAIL-M \\
        \bottomrule
    \end{tabular}
\end{table*}